\documentclass{article}

\usepackage{makecell}
\usepackage{PRIMEarxiv}
\usepackage{float} 
\usepackage[utf8]{inputenc} 
\usepackage[T1]{fontenc}    
\usepackage{hyperref}       
\usepackage{url}            
\usepackage{booktabs}       
\usepackage{multirow}
\usepackage{amsfonts}       
\usepackage{nicefrac}       
\usepackage{microtype}      
\usepackage{lipsum}
\usepackage{fancyhdr}       
\usepackage{graphicx}       
\graphicspath{{media/}}     
\usepackage{amsmath}
\title{Improving Phishing Email Detection Performance of Small Large Language Models}

\author{
    Zijie Lin\\
    National University of Singapore \\
    \texttt{lin.zijie@u.nus.edu} \\
    \And
    Zikang Liu\\
    National University of Singapore \\
    \texttt{e1351201@u.nus.edu} \\
    \And
    Hanbo Fan\\
    National University of Singapore \\
    \texttt{e1351318@u.nus.edu} \\
}

\begin{document}
\maketitle

\begin{abstract}
Large language models(LLMs) have demonstrated remarkable performance on many natural language processing(NLP) tasks and have been employed in phishing email detection research. However, in current studies, well-performing LLMs typically contain billions or even tens of billions of parameters, requiring enormous computational resources. To reduce computational costs, we investigated the effectiveness of small-parameter LLMs for phishing email detection. These LLMs have around 3 billion parameters and can run on consumer-grade GPUs. However, small LLMs often perform poorly in phishing email detection task. To address these issues, we designed a set of methods including Prompt Engineering, Explanation Augmented Fine-tuning, and Model Ensemble to improve phishing email detection capabilities of small LLMs. We validated the effectiveness of our approach through experiments, significantly improving both accuracy and F1 score on the SpamAssassin and CEAS\_08 datasets. Furthermore, the fine-tuned models demonstrated strong transferability, achieving robust performance across multiple unseen phishing datasets, outperforming traditional baselines and approaching standard-sized LLMs.
\end{abstract}

\keywords{Phishing Email Detection \and LLMs \and Fine-tuning}

\section{Introduction}
Phishing emails have long represented a significant cybersecurity threat, posing serious risks to both individuals and organizations. Various phishing detection techniques have been researched over time. Initially, rule-based filtering systems and simple blacklists were used for phishing email detection. Later, simple machine learning methods such as Naive Bayes \cite{Bayesian} and Support Vector Machine (SVM) \cite{SVM}, along with data mining approaches like decision trees \cite{DecisionTree}, marked significant developments in email detection technology.

In recent years, deep learning techniques have received increasing attention in phishing email detection. Deep neural network models can automatically capture high-dimensional features of emails and have been widely studied in phishing detection \cite{DLForPhishing}. Initially, researchers focused on using Convolutional Neural Networks (CNN)\cite{CNN}, Recurrent Neural Networks (RNN)\cite{RNN}, and similar models for phishing email detection. In recent years, Transformer-based models have developed rapidly\cite{Transformer}. These models excel at capturing semantic information and were quickly applied to phishing detection tasks. At first, people preferred Encoder-Only architectures like BERT\cite{BERT}, as its bidirectional attention mechanism helped better capture contextual information. As Decoder-Only models like GPT\cite{ChatGPT, GPT-4} increased in parameter count and training data quality, these LLMs also became effective at phishing email detection tasks.

\begin{figure}
  \centering
  \includegraphics[width=0.8\textwidth]{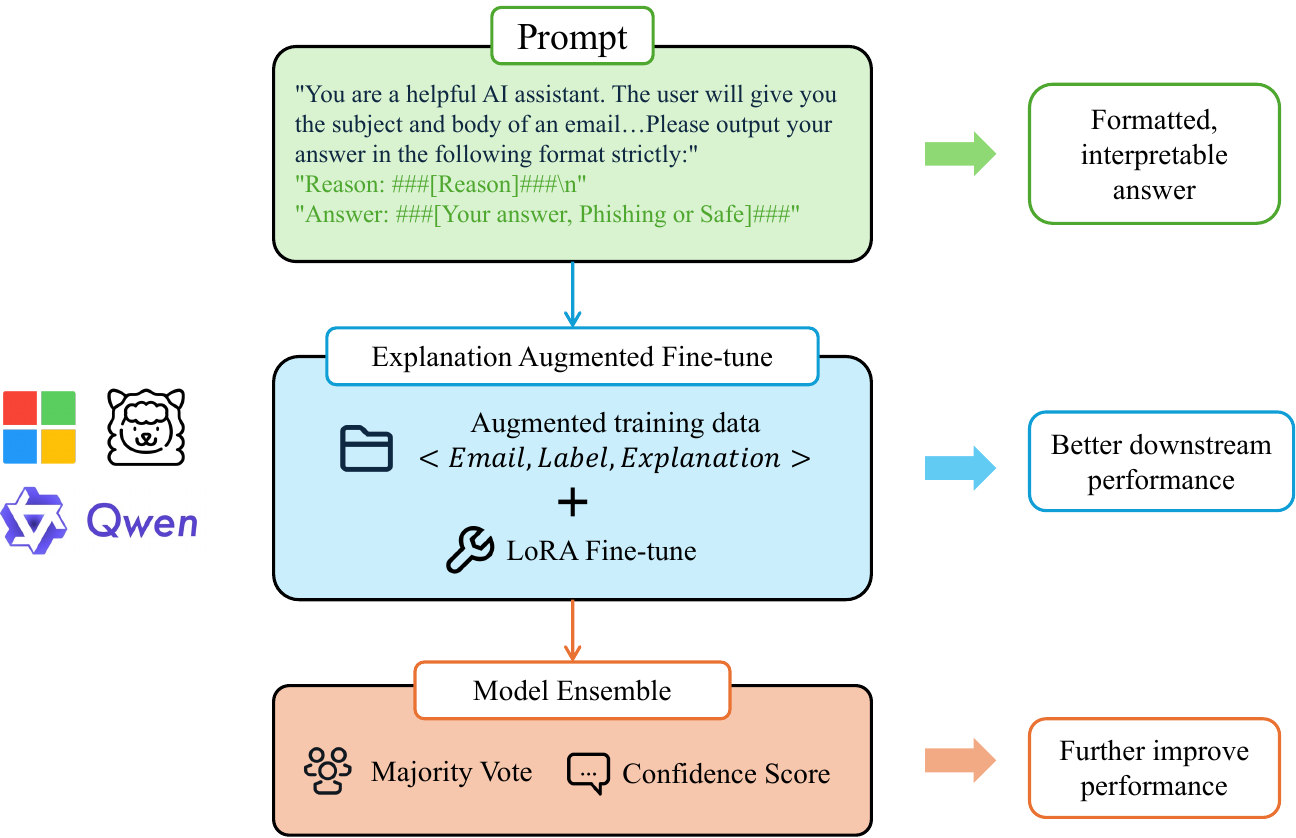}
  \caption{The workflow of improving phishing email detection performance of small LLMs}
  \label{fig:workflow}
\end{figure}

However, we note that although many studies proposed using LLMs for phishing detection, well-performing models typically come with larger parameter counts. For example, Ngoc Tuong Vy Nguyen et al. used a debate framework built with LLaMA-3.1-70B and GPT-4 for phishing detection\cite{DebatePhishing}, achieving approximately 0.99 accuracy on datasets like SpamAssassin and Nigerian Fraud\cite{Kaggle}. Despite their high accuracy, considering the hundreds of billions or even trillions of parameters in models like LLaMA-3.1-70B\cite{LLAMA3.1} and GPT-4\cite{GPT-4}, the computational costs are substantial. To address this issue, we propose using smaller language models such as LLaMA-3.2-3B-Instruct\cite{LLAMA3.2} and Phi-4-mini-Instruct\cite{Phi-4-mini} for detection. These models have around 3 billion parameters, can run on consumer-grade GPUs, and significantly reduce computational requirements. Furthermore, these smaller LLMs can provide text-level evidence for their judgments, thereby offering a degree of interpretability to the results. However, we found that directly using these models for detection yields poor results (e.g. strong bias or inconsistent output formats). To address these challenges, we propose several methods to enhance the performance of small-parameter LLMs in phishing email detection tasks. By practicing our methods step by step, we effectively improved the phishing email detection ability of small LLMs. The process of our method is shown in Figure \ref{fig:workflow}.

\begin{itemize}
\item \textbf{Prompt Engineering}: Through carefully designed prompts, we encourage LLMs to output both their reasoning and answer in a structured format, enabling basic phishing email detection.
\item \textbf{Explanation Augmented Fine-tuning}: Fine-tuning can enhance LLMs' capabilities on specific downstream tasks. However, we found that directly fine-tuning on datasets consisting only of emails and their labels yields poor results. We believe that using only email labels as training targets during fine-tuning doesn't align with the generative nature of LLMs. Therefore, we used GPT-4o-mini\cite{GPT-4o-mini} to generate an explanation for each training sample, detailing why the email is (or is not) a phishing attempt. Then, we fine-tuned each LLM on datasets consisting of emails, labels and explanations. Through this approach, we effectively improved the detection capabilities of fine-tuned LLMs. 
\item \textbf{Model Ensemble}: In the previous step, we fine-tuned multiple LLMs, allowing us to consider model ensemble techniques to further improve detection performance. We implemented two ensemble methods: (1) Majority Vote, where we have three LLMs evaluate each email and take the majority decision as the final result; (2) Confidence Ensemble, where we calculate confidence scores based on the token logprobs from LLM outputs and select the answer with the highest confidence score. Through model ensemble techniques, we found that it is possible to further enhance detection performance.

\end{itemize}
\section{Related Work}

There are significant content differences between phishing emails and legitimate emails. Early studies mainly utilized keyword filters to detect phishing emails. These filters are typically based on the TF-IDF algorithm or the bag-of-words model, combined with mutual information to extract keyword features. Androutsopoulos et al. evaluated the detection abilities of various types of Naive Bayes classifiers and achieved some results \cite{keyword_Naive_Bayes1, keyword_Naive_Bayes2}. However, their approaches are based on idealized assumptions and are insufficient for countering ever-evolving phishing threats. Drucker et al. employed SVM with a linear kernel and addressed data imbalance issues in real-world environments through weighted strategies. Their model demonstrated good performance in handling high-dimensional sparse features \cite{keyword_SVM}. Carreras et al. explored the performance of boosting tree models and the influence of base learners with varying complexities. Their model showed a higher accuracy than traditional linear models and could adapt to different misclassification costs \cite{keyword_Boosting_Tree}. 

The development of deep learning has introduced new opportunities for phishing email detection. Roy et al. utilized word embedding techniques to extract features. They were able to capture deep semantic information in emails. They trained CNN and LSTM models on these features, which significantly outperformed traditional models in phishing detection tasks \cite{deeplearning_CNN_LSTM}. Guo et al. employed a pre-trained BERT model to extract semantic features from email texts and trained four traditional classifiers—Naive Bayes, SVM, Logistic Regression, and Random Forest—to evaluate performance. Their results demonstrated that semantic features extracted by BERT could significantly enhance the effectiveness of traditional models \cite{deeplearning_Bert}.

With the continued advancement of deep learning, generative LLMs have emerged. These models exhibit exceptional capabilities in contextual understanding and reasoning, achieving outstanding performance in text classification tasks. Cunha et al. investigated the differences between traditional models and LLMs in the text classification task and demonstrated the significant superiority of LLMs \cite{LLMs_Text_Classification}. Heiding et al. further explored the effectiveness of LLMs in phishing email detection tasks \cite{LLMs_Phishing_email_1}. Their findings revealed that LLMs excel in identifying the intent behind phishing emails, even emails with ambiguous intent. In some scenarios, the LLMs even showed better performance than human evaluators. Koide et al. employed GPT-4 for phishing email detection, achieving an impressive accuracy of 0.997 \cite{LLMs_Phishing_email_2}. Through prompt engineering, they adapted LLMs to handle diverse phishing attack strategies. Moreover, prompt engineering could also improve interpretability, allowing the model to provide structured reasoning to users and increase the trust of users.

Despite the outstanding performance of LLMs in phishing email detection tasks, a critical challenge remains: the large number of parameters imposes significant obstacles to model deployment and costly API call expenses. Additionally, fine-tuning these large LLMs for downstream tasks demands substantial computational and time resources. Our study focuses on evaluating the performance of small LLMs in phishing email detection. Through fine-tuning, these smaller LLMs can achieve performance that meets or even surpasses that of large LLMs, making them an excellent choice for balancing cost and effectiveness while preserving interpretability.

\section{Method}

\subsection{Problem Statement}

Our study focuses on the detection of phishing email. Given a phishing email $E$, including its subject $S$ and body $B$, suppose we use model $M$ with parameters $\theta$ for identification. We expect the model to output a label $L$, where $L$ indicates whether the email is a phishing email or not:

\begin{equation*}
L = M(<S,B>|\theta)
\end{equation*}

When using LLMs for detection, $L$ is no longer a label, but rather a text judgment about whether the email is a phishing email, specifically \textit{Phishing} or \textit{Safe}. We also need to input a prompt $P$ to instruct the model to output content according to our requirements:

\begin{equation*}
L = M(<S,B>|\theta,P)
\end{equation*}

\subsection{Prompt Engineering}

After instruction fine-tuning, LLMs are capable of following human instruction. However, prompt design still significantly impacts LLMs' performance for specific tasks. We discovered that when simply asking LLMs to output the word \textit{Phishing} or \textit{Safe} as a label, they often fail to strictly adhere to the requested format, creating difficulties in extracting the model's decisions and reducing overall usability. Additionally, when instructed to output just a single word, LLMs exhibit notable biases, showing a strong tendency to favor either \textit{Phishing} or \textit{Safe} regardless of the input. We believe this occurs because requiring models to output only a label creates a disconnect from their pretraining objectives, as LLMs are inherently better suited for open-ended generative tasks. Therefore, we modified our prompting strategy to first have the LLM articulate its reasoning for the judgment, and only afterward output the final answer enclosed in special symbols, leveraging the model's natural strengths while ensuring consistent, extractable results.

\subsection{Explanation Augmented Fine-tuning}

\begin{figure}
  \centering
  \includegraphics[width=0.8\textwidth]{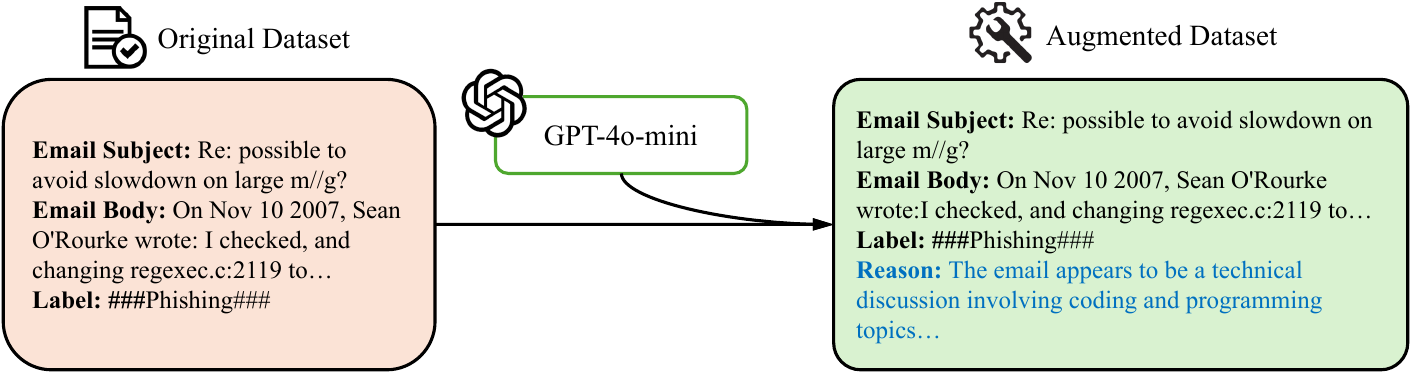}
  \caption{Using GPT-4o-mini to augment the original phishing email dataset with explanation}
  \label{fig:augment}
\end{figure}

The pretrain-finetune paradigm effectively improves LLMs' capabilities across various downstream tasks\cite{BERT, GPT}. We first considered directly fine-tuning small LLMs using phishing email datasets, incorporating email subjects $S$, body $B$, and actual labels $L$. However, we found that this approach yielded very limited improvements in phishing detection performance. We believe that while generative LLMs possess powerful open-ended text generation abilities and demonstrate strong generalization across various NLP tasks, requiring them to output a single word (either \textit{Phishing} or \textit{Safe}) resembles a closed-ended answer task that misaligns with their pretraining objectives. This deficiency persists when using single-word labels as pretraining targets. To address this issue, we expanded our fine-tuning data objectives by augmenting the training targets to include both explanations and labels. We input the emails and labels from our training data into the GPT-4o-mini model, as shown in Fig.~\ref{fig:augment}, instructing it to generate explanations for email classifications, thereby creating augmented training data. Each augmented training data item contains an email subject $S$, email body $B$, explanation $R$, and label $L$. Through explanation augmented fine-tuning, our training data more closely resembles open-ended generation tasks, better facilitating LLM fine-tuning. Additionally, by encouraging LLMs to provide explanations, we can mitigate hallucination issues, improving overall detection reliability and interpretability.

We then use the augmented training data to fine-tune small LLMs. We do not apply full parameter fine-tuning due to the limitation of computing resources. Instead, we decide to apply Low-Rank Adaptation (LoRA) to fine-tune the LLMs\cite{LoRA}. The core idea of LoRA is to decompose the weight matrix $W$ into two smaller matrices $A$ and $B$, such that:

\begin{equation}
    W^{'}=W+\Delta{W}=W+AB
\end{equation}

where $W$ is frozen and $\Delta{W}$ is the weight update during fine-tuning, $B \in \mathbb{R}^{d \times r}$, $A \in \mathbb{R}^{r \times k} $($r \ll \min(d,k)$). The final forward computation is scaled as: 
\begin{equation}
    h = Wx + \Delta Wx = Wx + \frac{\alpha}{r}ABx
\end{equation}

where $\alpha$ is an adjustable scaling factor. LoRA reduces memory requirements during fine-tuning by training only low-rank decomposition matrices ($A$ and $B$) instead of the full weight matrix W, significantly decreasing the number of parameters that need gradient storage and optimizer states.

\subsection{Model Ensemble}

Since different LLMs may possess varying knowledge and reasoning capabilities, we propose two model ensemble methods to achieve better phishing detection performance: (1) Confidence score ensemble, which calculates confidence scores based on the log probabilities of tokens output by LLMs, and selects the answer with the highest confidence as the final answer; (2) Majority vote, which selects the answer supported by the majority of models as the final answer.

In the forward pass, the final logits pass through a softmax layer to obtain probabilities corresponding to different tokens, which we can obtain in the form of logprobs. The sequence probability is calculated by exponentiating the logprobs of each token multiplying these probabilities together, and then taking the $N$th root of the result, where $N$ represents the sequence length. This process effectively computes the geometric mean of token probabilities, providing a Length-Normalized(LN) confidence score, see Equation \ref{equ:LNConf}.

Majority voting is rather straightforward. In our experiments, we fine-tuned three small LLMs: LLaMA-3.2-3B-Instruct, Phi-4-mini-Instruct, and Qwen-2.5-1.5B-Instruct. For a target email, if two or more models classify it as Phishing (or Safe), we adopt the majority opinion as the final result.

\begin{equation}
    \label{equ:LNConf}
    LN~Conf.=\left(\prod_{i=1}^{N} \exp(logprob_{i})\right)^{1/N}
\end{equation}

\section{Experiments}
In this section, we conduct experiments to demonstrate the effectiveness of our method. The small LLMs employed include LLaMA-3.2-3B-Instruct\cite{LLAMA3.2}, Phi-4-mini-Instruct\cite{Phi-4-mini}, and Qwen-2.5-1.5B-Instruct\cite{Qwen}, with parameters of 3 billion, 3.8 billion, and 1.5 billion, respectively. We fine-tuned the three LLMs separately on SpamAssassin and CEAS\_08. Each model was fine-tuned and evaluated separately on each of the two datasets. Specifically, for each dataset, a model was fine-tuned on its training split and subsequently evaluated on its corresponding test split. Both inference and fine-tuning can be performed on one RTX 3090 GPU, and we limit the length of training sequences and output sequences to avoid GPU memory overflow issues. We first conduct zero-shot experiments based solely on prompt engineering, followed by explanation-augmented fine-tuning and model ensemble experiments. We also perform ablation studies to illustrate the significant performance improvements achieved through explanation enhancement in the training data. Moreover, we evaluate the transferability of our fine-tuned LLMs compared with traditional models and standard-sized LLMs. There are 1,069 test samples in the SpamAssassin test set, and 1,000 test samples each in the CEAS\_08, Ling, and Enron test sets.

\textbf{Evaluation Metrics}: We report standard classification metrics including Precision, Recall, and F1 score, computed as:
\begin{equation*}
\text{Precision} = \frac{TP}{TP + FP}
\end{equation*}

\begin{equation*}
\text{Recall} = \frac{TP}{TP + FN}
\end{equation*}

\begin{equation*}
F1\text{-Score} = 2 \times \frac{\text{Precision} \times \text{Recall}}{\text{Precision} + \text{Recall}}
\end{equation*}

where TP, FP, FN denote the true positives, false positives, and false negatives, respectively.

\subsection{Vanilla Prompt Engineering Performance}

We selected small LLMs such as LLaMA, Phi, and others for our baseline experiments. Compared to LLMs like GPT-4 that contain hundreds of billions or even trillions of parameters, these small LLMs contain only around 3 billion parameters. Through prompting, we encouraged the LLMs to output their reasoning before making phishing email judgments, and to wrap their answers in special symbols (wrapped by \#\#\#). Table \ref{tab:1} presents our experimental results. If an email is a phishing email, it is labeled as a positive sample.

\begin{table} [h]
 \caption{The experimental results of Vanilla Prompting}
  \centering
  \begin{tabular}{l|cccc|cccc}
    \toprule
     \multirow{2}{*}{\centering\makecell[c]{Model}} & \multicolumn{4}{c|}{SpamAssassin} & \multicolumn{4}{c}{CEAS\_08} \\
    \cmidrule(r){2-5} \cmidrule(l){6-9}
    & Accuracy & F1 & Precision & Recall & Accuracy & F1 & Precision & Recall \\
    \midrule
    LLaMA-3.2-3B-Instruct   & 0.587 & 0.354 & 0.317 & 0.401 & 0.671 & 0.566 & 0.591 & 0.543 \\
    Phi-4-mini-Instruct              & 0.647 & 0.537 & 0.423 & 0.734 & 0.808 & 0.798 & 0.804 & 0.793 \\
    Qwen-2.5-1.5B-Instruct  & 0.388 & 0.469 & 0.899 & 0.317 & 0.657 & 0.758 & 0.641 & 0.927 \\
    \bottomrule
  \end{tabular}
  \label{tab:1}
\end{table}

From the results, we observe that applying Vanilla Prompting to small LLMs yields unsatisfactory performance. On the SpamAssassin dataset, accuracies are below 0.65 and F1-scores hover around 0.5. Although performance on CEAS\_08 is relatively better, the best accuracy and F1-score are only around 0.8. These results demonstrate that small LLMs without fine-tuning are inadequate for phishing email detection tasks (especially in high-risk environments), underscoring the need for enhancement methods such as fine-tuning.

\subsection{Explanation Augmented Fine-tuning Performance}

\begin{table}[h]
  \caption{The experimental results of GPT-3.5-Turbo, GPT-4o-mini, traditional ML models and EA(Explanation Augmented Fine-tuned) LLMs}
  \centering
  \begin{tabular}{l|cccc|cccc}
    \toprule
    \multirow{2}{*}{Model} & \multicolumn{4}{c|}{SpamAssassin} & \multicolumn{4}{c}{CEAS\_08} \\
    \cmidrule(r){2-5} \cmidrule(l){6-9}
    & Accuracy & F1 & Precision & Recall & Accuracy & F1 & Precision & Recall \\
    \midrule
    Naive Bayes            & 0.852 & 0.760 & 0.740 & 0.780 & 0.829 & 0.835 & 0.947 & 0.746 \\
    SVM                    & 0.967 & 0.945 & 0.950 & 0.940 & 0.989 & 0.991 & 0.990 & 0.991 \\
    XGBoost                & 0.953 & 0.920 & 0.941 & 0.900 & 0.997 & 0.997 & 0.998 & 0.997 \\
    GPT-3.5-Turbo          & 0.757 & 0.705 & 0.553 & 0.972 & 0.880 & 0.897 & 0.882 & 0.912 \\
    GPT-4o-mini            & 0.948 & 0.915 & 0.941 & 0.891 & 0.854 & 0.859 & 0.966 & 0.773 \\
    LLaMA-3.1-70B-Instruct & 0.801 & 0.635 & 0.693 & 0.586 & 0.881 & 0.896 & 0.882 & 0.910 \\
    \midrule
    LLaMA-3.2-3B-Instruct-EA   & 0.963 & 0.928 & 0.934 & 0.922 & 0.891 & 0.893 & 0.955 & 0.838 \\
    Phi-4-mini-Instruct-EA     & 0.968 & 0.944 & 0.925 & 0.963 & 0.917 & 0.927 & 0.932 & 0.922 \\
    Qwen-2.5-1.5B-Instruct-EA  & 0.860 & 0.673 & 0.863 & 0.551 & 0.860 & 0.868 & 0.907 & 0.833 \\
    \bottomrule
  \end{tabular}
  \label{tab:2}
\end{table}

We then perform augmented fine-tuning on selected LLMs using the SpamAssassin and CEAS\_08 datasets respectively. We extract a subset of 1,000 samples from the training set and perform explanation augmentation on this subset. We then fine-tune these LLMs with this training set using LoRA. We compare the results of our fine-tuned small LLMs with traditional ML models such as Naive Bayes, Support Vector Machine (SVM), and XGBoost, as along with standard-sized LLMs such as GPT-3.5-Turbo, GPT-4o-mini and LLaMA-3.1-70B-Instruct. When using traditional ML, we employed the paraphrase-MiniLM-L3-v2 embedding model from Sentence-BERT to embed email subject and body, and used the resulting embedding vectors for training and testing\cite{Sentence_Bert}. Unlike LLMs, traditional models are trained without explanation augmentation, because they are not designed for open-ended generative tasks. Thus, their training sets only include email content and labels, without any explanation. The results are shown in Table \ref{tab:2} .

After explanation augmented fine-tuning, the performance of small LLMs improved significantly. On the SpamAssassin dataset, the accuracy and F1 score of LLaMA-3.2-3B-Instruct model increased from 0.587 and 0.543 to 0.963 and 0.928, respectively, while Phi-4-mini-Instruct's increased from 0.647 and 0.537 to 0.968 and 0.944. Qwen-2.5-1.5B-Instruct's accuracy improved by 122\%, from 0.388 to 0.860. On the CEAS\_08 dataset, the LLaMA-3.2-3B-Instruct model achieved an accuracy of 0.891 and an F1 score of 0.893, while Phi-4-mini-Instruct reached 0.917 and 0.927, respectively. Although these results are marginally lower than their corresponding performance on the SpamAssassin dataset, they nonetheless demonstrate a substantial improvement over their vanilla baselines. Remarkably, Qwen-2.5-1.5B-Instruct attained an accuracy of 0.860 and an F1 score of 0.868, outperforming its performance on the SpamAssassin dataset. Although the performance of Qwen-2.5-1.5B was not as good as LLaMA-3.2-3B and Phi-4-mini-Instruct, due to its smaller parameter counts, it still showed substantial improvement over its own vanilla baseline. Additionally, these small LLMs even outperformed larger LLMs after fine-tuning. For example, on the SpamAssassin dataset, Phi-4-mini-Instruct's accuracy was 27.9\% higher than GPT-3.5-Turbo, 20.8\% higher than LLaMA-3.1-70B-Instruct, and 2.1\% higher than GPT-4o-mini. The fine-tuned LLaMA-3.2-3B-Instruct also achieved better accuracy and F1 scores than these three large LLMs. On the CEAS\_08 dataset, both Phi-4-mini-Instruct and LLaMA-3.2-3B-Instruct surpassed larger-scale LLMs. Among them, Phi-4-mini-Instruct delivered the most competitive results, achieving an accuracy 4.2\% higher than GPT-3.5-Turbo, 7.4\% higher than GPT-4o-mini, and 4.1\% higher than LLaMA-3.1-70B-Instruct. The performance of small LLMs was similar to XGBoost and SVM, but they can provide interpretable judgment results in the form of human language. 

Our experimental results demonstrate that after fine-tuning, small LLMs exhibit substantially enhanced phishing email detection capabilities, achieving performance metrics comparable to (in some cases exceeding) those of established machine learning approaches. Notably, these fine-tuned compact models outperform larger LLMs with orders of magnitude more parameters, highlighting the effectiveness of our approach in optimizing model efficiency for detecting phishing emails.

\subsection{Model Ensemble}

After fine-tuning multiple LLMs, we explored model ensemble strategies to potentially leverage the complementary strengths of different LLMs. We implemented two ensemble methods: confidence ensemble and majority vote. The comparative results between ensemble approaches and individual fine-tuned models are presented in Table \ref{tab:3}. For the confidence ensemble, we specifically utilized LLaMA-3.2-3B-Instruct and Phi-4-mini-Instruct, while the majority vote incorporated all three fine-tuned small LLMs.

The experimental results demonstrate that on the SpamAssassin dataset, both ensemble methods achieved notable improvements, with accuracies reaching approximately 0.975 and F1 scores of 0.953 and 0.959 for confidence ensemble and majority vote, respectively. However, on the CEAS\_08 dataset, neither ensemble method significantly outperformed the best individual model (Phi-4-mini-Instruct), suggesting that the ensemble benefits may vary across different datasets.

While model ensemble strategies demonstrated potential in enhancing the robustness and performance of LLMs in phishing email detection, the marginal improvements were relatively modest compared to the substantial gains achieved through explanation-augmented fine-tuning alone.

\begin{table}
 \caption{The experimental results of Confidence Ensemble and Majority Vote comparing to using EA(Explanation
Augmented Fine-tuned) small LLMs}
  \centering
  \begin{tabular}{l|cc|cc}
    \toprule
    \multirow{2}{*}{Model} & \multicolumn{2}{c}{SpamAssassin} & \multicolumn{2}{|c}{CEAS\_08} \\
    \cmidrule(lr){2-3} \cmidrule(lr){4-5}
    & Accuracy & F1 & Accuracy & F1  \\
    \midrule
    LLaMA-3.2-3B-Instruct-EA   & 0.963 & 0.928 & 0.891 & 0.893 \\
    Phi-4-mini-Instruct-EA     & 0.968 & 0.944 & 0.917 & 0.927 \\
    Qwen-2.5-1.5B-Instruct-EA  & 0.860 & 0.673 & 0.860 & 0.868 \\
    \midrule
    Confidence Ensemble        & 0.975 & 0.953 & 0.895 & 0.906 \\
    Majority Vote              & 0.976 & 0.959 & 0.884 & 0.896 \\
    \bottomrule
  \end{tabular}
  \label{tab:3}
\end{table}

\subsection{Ablation Study}

In this section, we will discuss the effectiveness of explanation augmentation on data for improving the capability of small LLMs in detecting phishing emails. Our core idea is to enhance the fine-tuning data by adding explanations for email classification, thereby making the pattern of fine-tuning data more similar to the open text generation pattern that LLMs are adapted to. To validate our approach, we directly fine-tuned LLMs using data without explanation augmentation and conducted experiments on the test set. Our experimental results are shown in Table \ref{tab:4}.

Our experiments reveal that without explanations in the fine-tuning data, the final prediction results are significantly compromised. The accuracy of the LLaMA-3.2-3B-Instruct model decreased by 40.7\%, while its F1 score plummeted from 0.928 to 0.219. The accuracy and F1 score of Qwen-2.5-1.5B-Instruct decreased by 36.4\% and 26.0\%, respectively. Phi-4-mini-Instruct demonstrated relatively robust performance, but its accuracy and F1 score still decreased by 15.1\% and 27.1\%, respectively. Overall, through ablation studies, we found that explanation augmentation has a significant impact on fine-tuning effectiveness, as it transforms our phishing email detection dataset from a closed-form prediction dataset to an open-ended generative dataset.

\begin{table}
 \caption{The experimental results of using and not using EA(Explanation Augmented Fine-tuned) small LLMs}
  \centering
  \begin{tabular}{l|cccc|cccc}
    \toprule
    \multirow{2}{*}{Model} & \multicolumn{4}{c|}{SpamAssassin} & \multicolumn{4}{c}{CEAS\_08} \\
    \cmidrule(r){2-5} \cmidrule(l){6-9}
    & Accuracy & F1 & Precision & Recall & Accuracy & F1 & Precision & Recall \\
    \midrule
    LLaMA-3.2-3B-Instruct-EA   & 0.963 & 0.928 & 0.934 & 0.922 & 0.891 & 0.893 & 0.955 & 0.838 \\
    Phi-4-mini-Instruct-EA     & 0.968 & 0.944 & 0.925 & 0.963 & 0.917 & 0.927 & 0.932 & 0.922 \\
    Qwen-2.5-1.5B-Instruct-EA  & 0.860 & 0.673 & 0.863 & 0.551 & 0.860 & 0.868 & 0.907 & 0.833 \\
    \midrule
    LLaMA-3.2-3B-Instruct-no-EA   & 0.571 & 0.219 & 0.221 & 0.216 & 0.346 & 0.147 & 0.158 & 0.137 \\
    Phi-4-mini-Instruct-no-EA     & 0.822 & 0.688 & 0.621 & 0.771 & 0.814 & 0.799 & 0.760 & 0.842 \\
    Qwen-2.5-1.5B-Instruct-no-EA  & 0.547 & 0.498 & 0.376 & 0.735 & 0.573 & 0.540 & 0.545 & 0.534 \\
    \bottomrule
  \end{tabular}
  \label{tab:4}
\end{table}

\subsection{Experiment on Transferability}
\begin{figure}[H]
  \centering
  \includegraphics[width=1.0\textwidth]{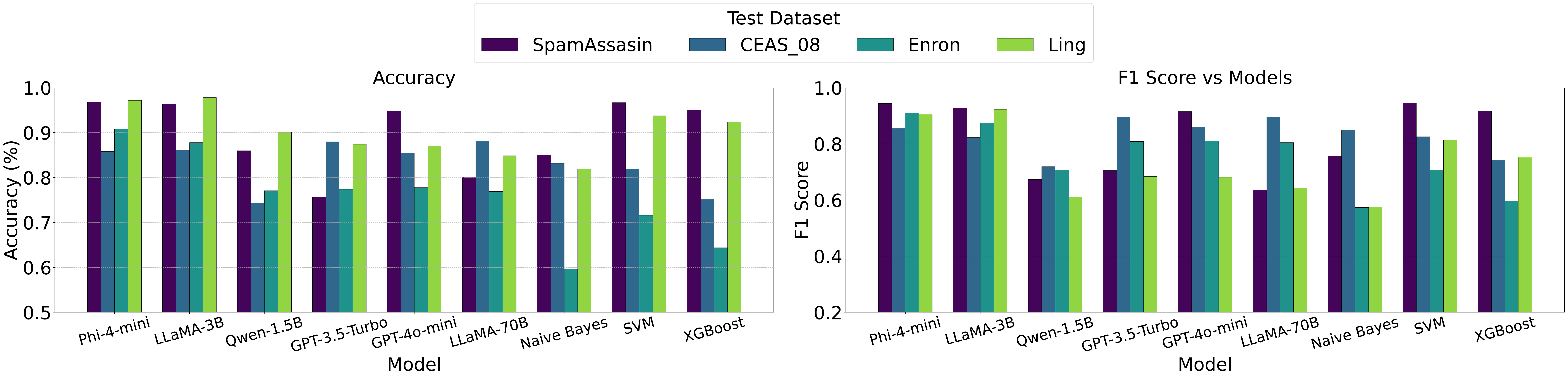}
  \caption{Accuracy and F1 scores of different models evaluated on multiple datasets. Note that small-scale LLMs and traditional machine learning models are specifically trained on the SpamAssassin dataset, while for standard-sized LLMs (e.g., GPT-3.5-Turbo), we employ direct prompting for phishing email detection without additional training.}
  \label{fig:workflow(EA_on_SpamAssasin)}
\end{figure}

\begin{figure}[H]
  \centering
  \includegraphics[width=1.0\textwidth]{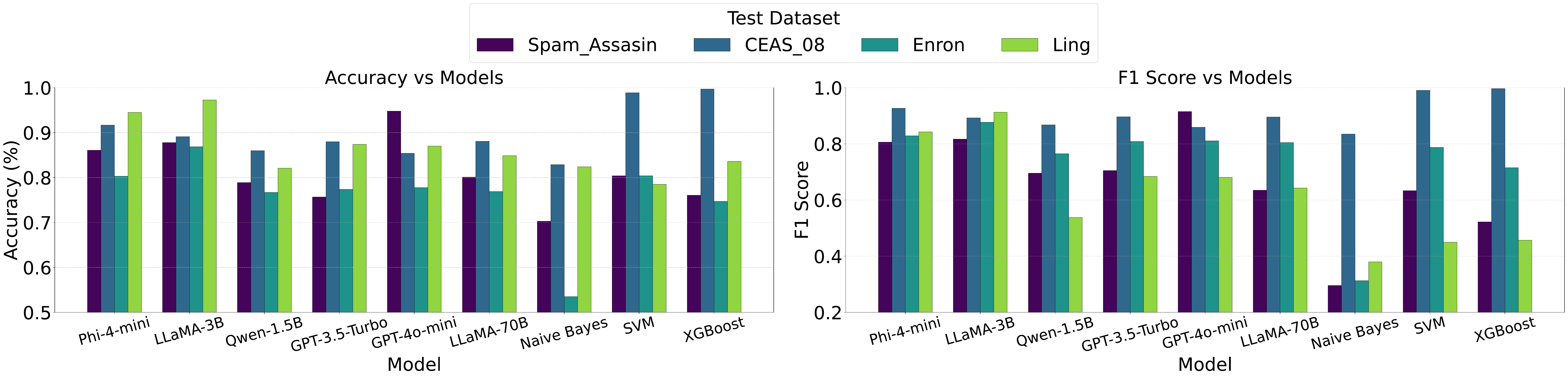}
  \caption{Accuracy and F1 scores of different models evaluated on multiple datasets. Note that small-scale LLMs and traditional machine learning models are specifically trained on the CEAS\_08 dataset, while for standard-sized LLMs (e.g., GPT-3.5-Turbo), we employ direct prompting for phishing email detection without additional training.}
  \label{fig:workflow(EA_on_CEAS08)}
\end{figure}

This section evaluates the transferability of our fine-tuned models alongside other methods. To simulate real-world generalization challenges, we trained small LLMs and traditional machine learning models (such as SVM, XGBoost and Naive Bayes) on a source dataset (either CEAS\_08 or SpamAssassin) and subsequently tested their performance on unseen datasets: Enron and Ling\cite{Kaggle}. For broader comparison, we also include the results of large LLMs like GPT-3.5-Turbo. We report the accuracy and F1 score to compare how well each model generalizes across domains as shown in Fig.~\ref{fig:workflow(EA_on_SpamAssasin)} and 
Fig.~\ref{fig:workflow(EA_on_CEAS08)}.

\textbf{Trained on SpamAssassin:} After fine-tuning on the SpamAssassin dataset, both small LLMs and traditional models (such as SVM) demonstrated excellent performance on this dataset. Except for the Qwen-2.5-1.5B-Instruct model, all models achieved approximately 0.95 in both accuracy and F1 scores, with Phi-4-mini-Instruct and LLaMA-3.2-3B-Instruct slightly outperforming traditional models like SVM. Small LLMs and traditional models also exhibited outstanding performance on the Ling dataset. Despite not being trained on the corresponding datasets, the explanation-enhanced fine-tuned Phi-4-mini-Instruct and LLaMA-3.2-3B-Instruct models achieved accuracies of 0.972 and 0.978, and F1 scores of 0.906 and 0.923, respectively. SVM and XGBoost models also attained high accuracies of 0.938 and 0.924, with F1 scores reaching 0.815 and 0.753.

However, on the CEAS\_08 and Enron datasets, small LLMs demonstrated notably better transferability compared to traditional models. Phi-4-mini-Instruct maintained an accuracy of 0.858 on the CEAS\_08 dataset, surpassing SVM and XGBoost by 4.8\% and 14.1\% respectively, while also achieving higher F1 scores. On the Enron dataset, Phi-4-mini-Instruct, LLaMA-3.2-3B-Instruct, and Qwen-2.5-1.5B-Instruct achieved accuracies of 0.908, 0.878, and 0.771 respectively, outperforming Naive Bayes, SVM, and XGBoost which scored 0.597, 0.716, and 0.644. Small LLMs generally achieved higher F1 scores as well.

Additionally, it is noteworthy that Phi-4-mini-Instruct and LLaMA-3.2-3B-Instruct outperformed the three regular-sized LLMs on all datasets except CEAS\_08, although the number of parameters of the latter is several times or even dozens of times that of the former.

\textbf{Trained on CEAS\_08:} After fine-tuning on the CEAS\_08 dataset, traditional models such as SVM and XGBoost demonstrated strong performance on the test set, achieving accuracy and F1 scores close to 1. However, their performance dropped dramatically on the other three datasets, with accuracies of only 0.785 and 0.836 on the Ling dataset, and F1 scores below 0.5. The highest accuracy these two models achieved on the remaining datasets was merely 0.804. In contrast, Phi-4-mini-Instruct and LLaMA-3.2-3B-Instruct models maintained accuracies above 0.85 and F1 scores above 0.8 across all four datasets, showing no significant dataset bias. Furthermore, regular-sized LLMs generally performed slightly worse than the fine-tuned small LLMs in most cases.

In summary, small LLMs enhanced through explanation augmented fine-tuning demonstrate superior performance not only on the training dataset but also across various unseen datasets. This empirical evidence strongly indicates that our approach exhibits better transferability compared to traditional machine learning models, while consistently outperforming regular-sized LLMs in detecting phishing emails.

\section{Limitation}

Our method effectively improves the ability of small LLMs in phishing email detection tasks, achieving performance exceeding models like LLaMA-3.1-70B-Instruct on models around 3B in size. However, due to time and experimental resource constraints, our research still has the following limitations:

\begin{itemize}

\item \textbf{Dataset:} We fine-tuned the three LLMs only on SpamAssassin and CEAS\_08 datasets. Additional experiments on other datasets could be added to further validate the effectiveness of our method.

\item \textbf{Transferability:} While we examined the transferability of our method, we did not further propose methods to improve the model's performance across various datasets.

\item \textbf{Quantifiable Cost Comparison:} Although it is intuitively less expensive to use small LLMs for inference, we did not specifically quantify the cost savings from using small LLMs for detection.

\item \textbf{Ensemble Strategy:} Beyond confidence ensemble and simple majority voting, more advanced methods, such as weighted voting based on model confidence or validation performance, may further improve performance but were not explored in this work.

\end{itemize}

\section{Conclusion}

In this study, we explore the feasibility of using small LLMs for phishing email detection. While directly using these small LLMs yields poor detection performance, we designed a process to improve their performance on the phishing detection task. By using prompt engineering to standardize the model output format and encouraging the model to output the reasoning behind its analysis, we first obtain formatted answers with explanations. Subsequently, we significantly improved the phishing email detection capabilities of the small LLMs through explanation augmented fine-tuning, achieving performance that matches or even exceeds mainstream machine learning methods and LLMs with tens of times more parameters. Furthermore, we found that it is possible to further improve the detection performance through model ensemble. Finally, our experiments show that fine-tuned small LLMs retain strong performance across multiple unseen datasets, outperforming traditional models and approaching standard-sized LLMs in terms of transferability. In summary, our research indicates the possibility of efficiently performing phishing email detection tasks using small LLMs, and we can obtain high-performance and explainable detection results from these small LLMs.

\bibliographystyle{unsrt}  
\bibliography{references}

\end{document}